\documentclass[letterpaper, 10 pt, conference]{ieeeconf} 
\usepackage{amsmath,amsfonts}
\usepackage{algorithmic}
\usepackage{array}
\usepackage[caption=false,font=normalsize,labelfont=sf,textfont=sf]{subfig}
\usepackage{textcomp}
\usepackage{stfloats}
\usepackage{url}
\usepackage{verbatim}
\usepackage{graphicx}

\usepackage{amsmath, amssymb}
\usepackage{makecell}
\usepackage{multirow}
\usepackage{algorithm}  
\usepackage{xcolor}         % colors
\usepackage{amsmath}  
\usepackage{xspace}

\IEEEoverridecommandlockouts
\renewcommand{\algorithmicrequire}{\textbf{Input:}}  % Use Input in the format of Algorithm  
\renewcommand{\algorithmicensure}{\textbf{Output:}} % Use Output in the format of Algorithm  

\def\BibTeX{{\rm B\kern-.05em{\sc i\kern-.025em b}\kern-.08em
    T\kern-.1667em\lower.7ex\hbox{E}\kern-.125emX}}
\usepackage{balance}
\usepackage{cite}
\usepackage[justification=centering]{caption}
\usepackage{booktabs}

\newcommand{\reflabel}{dummy} % Dummy initial reflabel - use renewcommand...

\newcommand{\be}{\begin{equation}}
\newcommand{\ee}{\end{equation}}
\newcommand{\eqlabel}[1]{\label{eq:\reflabel-#1}}
\renewcommand{\eqref}[2][\reflabel]{(\ref{eq:#1-#2})}

\newcommand{\seclabel}[1]{\label{sec:\reflabel-#1}}

\newcommand{\figlabel}[2][\reflabel]{\label{fig:#1-#2}}
\newcommand{\figref}[2][\reflabel]{Fig.~\ref{fig:#1-#2}}

\newcommand{\tablelabel}[2][\reflabel]{\label{table:#1-#2}}
\newcommand{\tableref}[2][\reflabel]{Table~\ref{table:#1-#2}}

\newcommand{\ie}{i.e.\xspace}
\newcommand{\eg}{e.g.\xspace}
\newcommand{\etc}{etc.\xspace}
\newcommand{\etal}{et al.\xspace}

\newcommand{\point}{\textbf{x}}
\newcommand{\dir}{\textbf{d}}
\newcommand{\nerf}{\mathbf{F_\theta}}
\newcommand{\mean}{\mu}
\newcommand{\uncert}{\sigma}
\newcommand{\numrayset}{R}

\newcommand{\g}{\mathbf{g_\phi}}

\begin{document}

%%%%%%%%% Title
\title{Efficient View Path Planning for Autonomous Implicit Reconstruction}
\author{Jing Zeng$^{1}$ Yanxu Li$^{1}$ Yunlong Ran$^1$ Shuo Li$^1$  Fei Gao$^1$ Lincheng Li$^2$ Shibo He$^1$ Jiming chen$^1$  Qi Ye$^1$ % <-this % stops a space
\thanks{$^1$State Key Laboratory of Industrial Control Technology, Zhejiang University, Hangzhou 310027, China. (Corresponding author: \emph{Qi Ye})
}%
\thanks{$^{2}$ Fuxi AI Lab, NetEase, Hangzhou, 310052, China}%
\thanks{E-mail: \{zengjing, qi.ye\}@zju.edu.cn}
}
\maketitle

%%%%%%%%% ABSTRACT

\begin{abstract}

Implicit neural representations have shown promising potential for the 3D scene reconstruction. Recent work applies it to autonomous 3D reconstruction by learning information gain for view path planning. Effective as it is, the computation of the information gain is expensive, and compared with that using volumetric representations, collision checking using the implicit representation for a 3D point is much slower. In the paper, we propose to 1) leverage a neural network as an implicit function approximator for the information gain field and 2) combine the implicit fine-grained representation with coarse volumetric representations to improve efficiency. 
Further with the improved efficiency, we propose a novel informative path planning based on a graph-based planner. Our method demonstrates significant improvements in the reconstruction quality and planning efficiency compared with autonomous reconstructions with implicit and explicit representations. We deploy the method on a real UAV and the results show that our method can plan informative views and reconstruct a scene with high quality.

\end{abstract}

% %%%%%%%%% BODY TEXT
\section{Introduction}

In this paper we study the problem of view path planning for a mobile robot to reconstruct high quality 3D models for priori unknown scenes. The robot is required to autonomously plan and execute a path that maximize the quality of the reconstructed 3D scenes. The autonomous reconstruction has wide applications, such as virtual reality, digital twin, autonomous driving, smart city~\etc.

Recently, implicit neural representations have shown compelling results in 3D scene reconstruction ~\cite{barron2022mip,tancik2022block,muller2022instant} and promising potentials in SLAM~\cite{sucar2021imap,zhu2022nice}. Ran~\etal~\cite{ran2022neurar} applies an implicit representation, represented by a multilayer perceptron (MLP) $\nerf$, to autonomous 3D reconstruction by learning information gain for the view path planning. Effective as it is in reconstructing fine-grained 3D scenes with high fidelity, calculating the information gain field for different viewpoints from the representation is inefficient:
multiple rays ($R$ rays) are required to cast through the scene and multiple points ($N$ points) are sampled on each ray to integrate the reconstruction uncertainty for the frustum covered by each viewpoint; all these points are passed to the MLP $\nerf$ to estimate the uncertainties. $R$ scales with the image size and $N$ scales with the scene size and the level of details required for the reconstruction quality. The other limitation is the inefficiency in the collision checking. To check a 3D point is free or not, the point is fed into $\nerf$ to get the density value while in explicit volumetric representations, only the memory for the point is queried. 

% On the other hand, in previous view path planning work~\cite{schmid2020efficient,kompis2021informed}, the ray casting and the integration to get the information gain for volumetric representations is also computational intensive as similar to NeurAR~\cite{ran2022neurar}. Constrained by the computational budget, sampling-based methods \eg RRT or RRT* are used as they calculate the information gain for a fewer number of points compared with graph-based methods \eg the A* algorithm. However, the path planned by the sampling-based methods is usually not guaranteed to be the shortest and exhibits zig-zagging effects. 

On the other hand, as the computation of the information gain for viewpoints is expensive, sampling-based methods \eg RRT or RRT* are preferred in previous view path planning work~\cite{schmid2020efficient,kompis2021informed,song2018surface,selin2019efficient}  as they typically require fewer number of sampled viewpoints, though many of them are not guaranteed to be optimal.

% and then less computation of the information gain compared with graph-based methods \eg the A* algorithm. However, the path planned by the sampling-based methods is usually not guaranteed to be the shortest. 
\begin{figure}[t]
    \centering
    \includegraphics[width=0.9\linewidth]{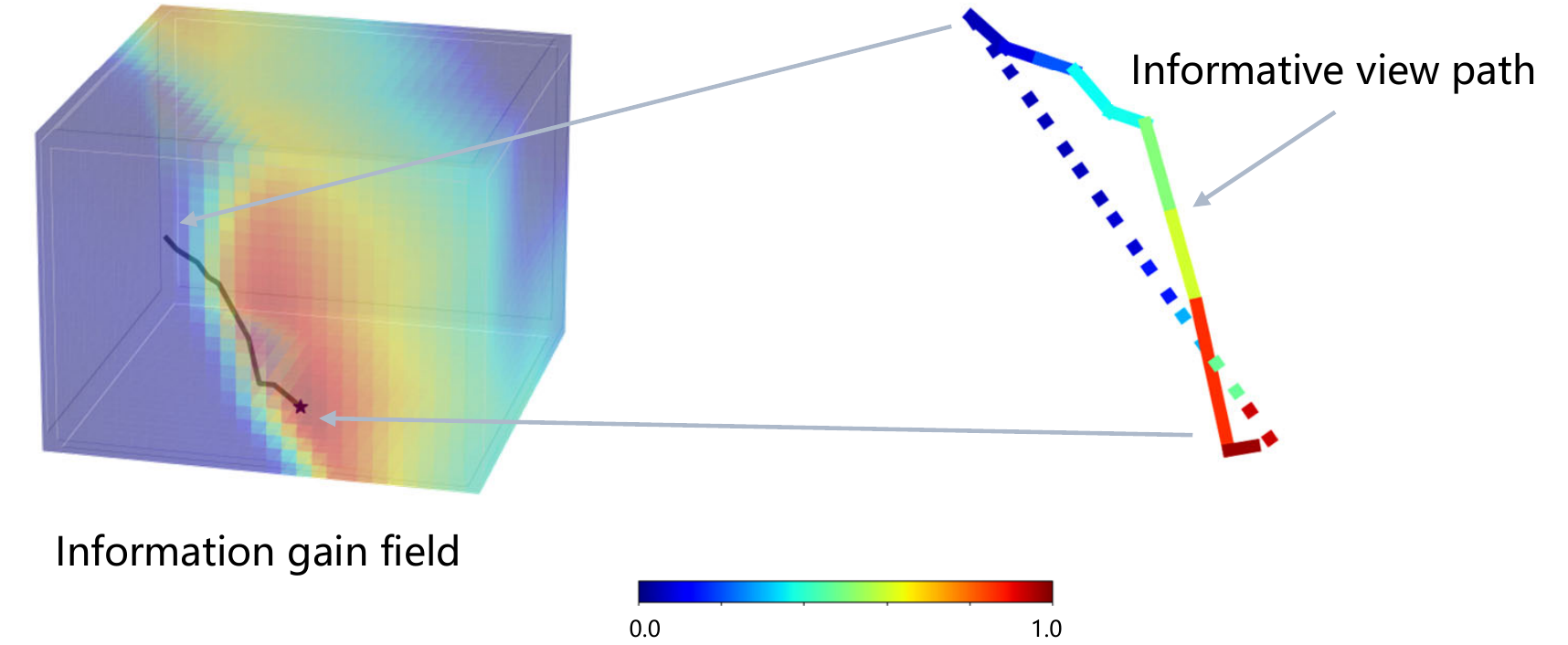}
\caption{Left: An information gain approximator $\g$ fitted for a local gain field  at the fifth step of the autonomous implicit reconstruction for the $cabin$ scene. Right: an informative view path planned by our method (the solid line) and the shortest path planned by A* (the dashed line). The path is colored  with the information gain. }
\figlabel{ig_filed}
 \vspace{-5mm}
\end{figure}

% \vspace{-5mm}

To address the above limitations, we propose an efficient view path planning for the autonomous reconstruction based on the novel implicit representations. Firstly, we assume the information gain field to be a smooth continuous function of the viewpoints and in the same spirit of the implicit scene representation, we propose to leverage a MLP $\g$ as an approximator for the field. With the approximation, getting the information gain of a viewpoint requires only one query of $\g$, instead of querying $\nerf$ for the scene radiance field for about a million times. Secondly, volumetric representations are introduced to complement to the implicit representations: the viewpoints to be sampled for the fitting of $\g$ are filtered based on a coarse TSDF to further reduce the computation of the information gain; an occupancy map from the TSDF is built for the fast collision checking. Thirdly, with the improved efficiency, the query of the information gain for a viewpoint is reduced to less than a millisecond, which opens the possibility of using planners providing almost optimal paths but requiring dense queries of the space for view path planning. Therefore, we demonstrate the possibility with a A* planner and propose a novel view path cost based on it.

To summarize, our contributions are:
\begin{itemize}
    \item We propose an implicit function approximator for the information gain field, which reduces the time complexity of querying the information gain for a viewpoint by at least $R\times N$ times.
    \item We propose a combination of implicit representations for fine-grained 3D scene reconstruction and volumetric representations for fast collision checking and viewpoints filtering.
    \item We propose a novel view path cost based on a graph based planner, which plans the shorter view paths and provides the better reconstruction quality than existing sampling-based planners.
\end{itemize}

\section{Related work}

The problem of determining the optimal viewpoints and paths for efficiently reconstructing a scene is known as the active vision or view-path-planning problem. This problem has been extensively studied for two decades~\cite{scott2003view,bircher2016receding,kompis2021informed,hardouin2020next,song2020active}. The most popular methods to solve the problem are frontier-based~\cite{liu2018floornet,selin2019efficient} and sampling-based methods~\cite{song2018surface,schmid2020efficient}. Frontier-based methods focus on the boundaries between known and unknown space to complete fast and global exploration tasks but are difficult to adapt to other tasks. Sampling-based methods focus on a Next-Best-View (NBV) strategy which selects NBV using feedback from the current partial reconstruction. NBV is determined using information gains of viewpoints which is defined over volumetric representations ~\cite{keselman2017intel,isler2016information,mendez2017taking} or surface-based representations~\cite{wu2014quality,schmid2020efficient,song2018surface}.

With the information gains of viewpoints defined, sampling-based planners are most commonly used to plan an informative view path. Rapidly Exploring Random Tree (RRT) is exploited to randomly sample viewpoints as nodes in RRT and use edges to connect viewpoints~\cite{mendez2017taking,schmid2020efficient,kompis2021informed}. Mendez et al.~\cite{mendez2017taking} use Sequential Monte Carlo to assign sampling weights in different locations for the information gain. A node with the maximum information gain is selected as the target node and the robot is guided along the tree to it.The method requires many sampling viewpoints for the convergence of SMC.  
Schmid et al.~\cite{schmid2020efficient} instead keep the entire tree alive and rewire the RRT after every planning iteration to reduce unnecessary re-computation of information gain. However, the rewiring process is time-consuming.
% and it is hard to find the optimal viewpoint in the tree. 
Kompis et al.~\cite{kompis2021informed} use an artificial potential field to predict the value  of proposed viewpoints to save the computation time of calculating their actual information gain. However, finding the frontiers of the surface is not efficient and predicting the value of viewpoints by manually defined parameters is not robust.

Due to the computational inefficiency of the information gain, planning methods sample a small subset of the viewpoints and focus on the design of reducing the number of sampling points for the information gain query. Our method, instead, focuses on improving the computation efficiency of the information gain. The super efficient querying of the information gain field then opens the possibility of investigating graph-based planning methods or other methods, which gives better optimality but requires the dense queries of the space for the view path planning problem.

% The approaches above do not necessarily find optimal viewpoints, spend much time on view sampling and evaluating. RRT-based methods plan a sub-optimal path with a longer length.  Therefore, we propose a view path planning methods based on sparse TSDF and NeRF, which use TSDF uncertainty to preselect view direction, define view information gain with TSDF distance and neural uncertainty, fit distribution of information gain with MLP, and plan an informative path for efficiently autonomous reconstruction  with A*.

\section{Method}

\seclabel{problem}
The problem considered is to generate a trajectory for a robot that yields high-quality 3D models of a bounded target scene and fulfills robot constraints like time and path length. The trajectory is composed of a sequence of paths and viewpoints $\Omega = (\omega_1...,\omega_n)$ where $\omega_i \in \mathcal{R}^3 \times \mathcal{SO}(2)$. Finding the best sequence of viewpoints $V$ is time-consuming and prohibitively expensive. Similar to~\cite{ran2022neurar,schmid2020efficient}, we adopt the greedy strategy and trade it as a Next-Best-View (NBV) problem. At each step of the reconstruction, the information gain of viewpoints within a certain range of the current position of a robot are evaluated based on the partial reconstruction scene. An informative path considering both the information gain and the path length is then planned and executed. Images captured along the view path are selected and are fed into the 3D reconstruction of next step. The process is repeated until some critera are met.

% \noindent\textbf{Limitations in Neural Uncertainty} Though NeurAR has proved the linear relationship between Neural Uncertainty and image quality PSNR, which guides view planning for 3D reconstruction. However, there are three problems not solved. Firstly, it defines the sample space as a spherical ring without obstacle avoidance. All viewpoints sampled are looking toward the center of the scene with a 20-degree yaw angle bias that is hard to find optimal view direction at each location. Secondly, Neural uncertainty only considers the color uncertainty of image pixels and doesn't include spatial depth information, which makes the uncertainty invalid when viewpoints are too near or far from the surface of the scene without spherical ring constraint. But the spherical ring is a very strong prior, so NeurAR is not robust and generalizable. Finally, the time cost in calculating Neural uncertainty is expensive which results in inefficient view path planning when sampling viewpoint uncertainty. Simultaneously, it ‘s hard to approach optimal viewpoint quality and path cost by RRT as sampling-based methods are easy to fall into local optimum\cite{schmid2020efficient}. 

\begin{figure}[htbp]
    \centering
    \includegraphics[width=1.0\linewidth]{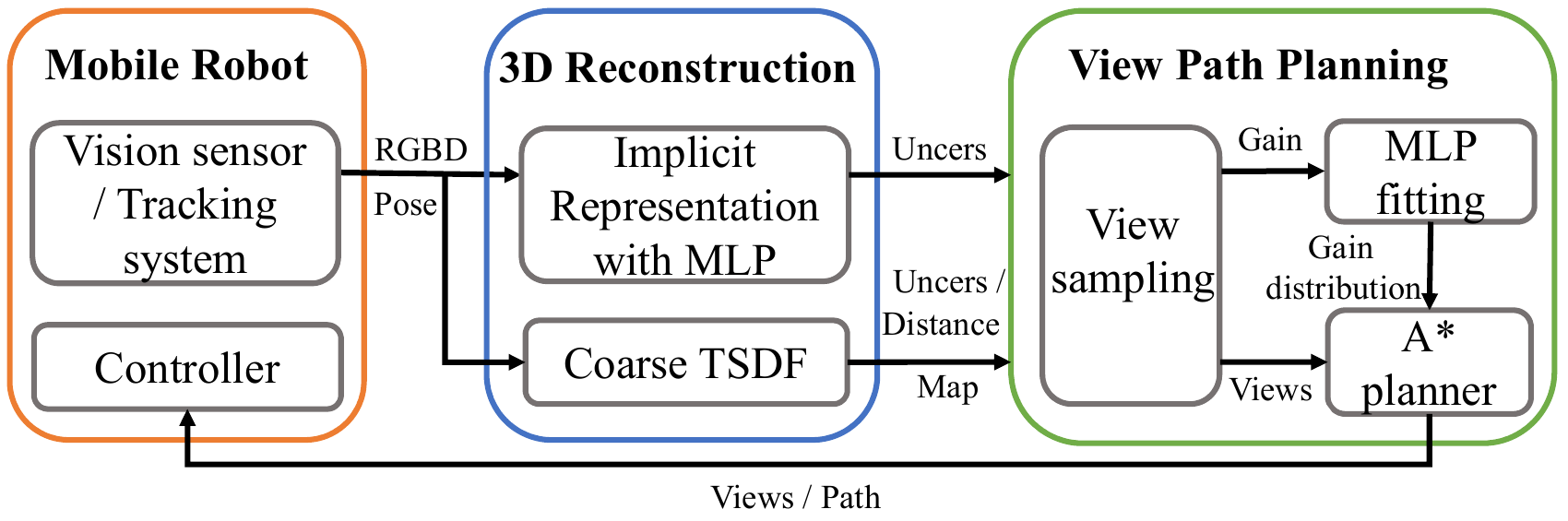}
\caption{The pipeline of our proposed method.} 
\figlabel{pipeline}
\vspace{-5mm}
\end{figure}

% \noindent\textbf{System Overview} Under the greedy strategy, our pipeline consists of three components: a Mobile Robot module, a 3D Reconstruction module and a View Path Planning module. 
% The Mobile Robot module takes the pictures at given viewpoints and the robot locates by a motion capture system. During the simulation, Unity Engine renders images at given viewpoints. 
% The 3D Reconstruction module reconstructs a scene by combining an implicit neural representation (\eg NeRF~\cite{}) and a volumetric representation (Coarse TSDF). The implicit neural representation provides high quality 3D models with fine grained details and also neural uncertainty (\secref{background}) as the information gain for view path planning. Coarse TSDF establishes an occupancy grid map for efficient distance and occupancy query, and viewpoint preselection. (\secref{coarse_tsdf}). 
% The View Path Planning module samples viewpoints from empty space, preselects view directions based on the coarse TSDF, and evaluates the information gain of viewpoints with TSDF surface distance and neural uncertainty (\secref{coarse_tsdf}, \secref{planner}). 

\subsection{System Overview} 

Under the greedy strategy, our pipeline consists of three components: a Mobile Robot module, a 3D Reconstruction module and a View Path Planning module in \figref{pipeline}. 
The Mobile Robot module takes the images at given viewpoints and the robot locates itself by a motion capture system. During the simulation, Unity Engine renders images at given viewpoints. 
The 3D Reconstruction module reconstructs a scene by combining an implicit neural representation (\eg NeRF~\cite{mildenhall2020nerf}) and a volumetric representation (A coarse TSDF). The implicit neural representation provides high quality 3D models with fine-grained details and also neural uncertainty as the information gain for the view path planning. The coarse TSDF filter viewpoints and establishes an occupancy grid map for efficient distance and occupancy query, and viewpoint filtering. 
The View Path Planning module first leverages volumetric representations for efficient viewpoint selection, approximates the information gain field by a MLP $\g$ and plans an informative view path based the A* algorithm.

% In the following section, we first briefly introduce the background about the implicit representation and its application in autonomous reconstruction in~\secref{background}. Then we elaborate our contributions regarding efficient view path planning: the approximation of the information gain field by a MLP and a combination of volumetric and implicit representation for efficient viewpoint sampling in~\secref{gainapprox}, and a novel view path planning algorithm balancing the reconstruction quality and path cost based on the A* planning~\secref{Aplan}.

\subsection{Background of Autonomous Implicit Reconstruction}
\seclabel{background}

Implicit representations has shown compelling results in recent years~\cite{sucar2021imap,martin2021nerf,zhu2022nice}. The representation fitting a scene by an implicit function. This function takes the direction of a ray $\dir$ and the location of a point $\point$ on the ray as inputs. Its outputs are the color value $c$ and density $\rho$ of the point. This function can be expressed as $\nerf(\point, \dir)=(c, \rho) $ and is implemented by a MLP. To deploy the implicit representation for autonomous reconstruction, NeurAR~\cite{ran2022neurar} learns neural uncertainty as the information gain. 
% It treats the color for a 3D point as a random variable under a Gaussian distribution $\mathcal{N} (\mean, \uncert^2) $ where the RGB channels share the same $\uncert$. The variance $\uncert$ quantifies the uncertainty of the reconstruction and composes the information gain. 
It expresses the scene function as $\nerf(\point, \dir)=( \mean, \uncert, \rho)$ and yields the neural uncertainty for a viewpoint 

% To deploy the implicit representation for autonomous reconstruction, NeurAR~\cite{ran2022neurar} learns neural uncertainty as the information gain. It treats the color for a 3D point as a random variable under a Gaussian distribution. The variance quantifies the uncertainty of the reconstruction and composes the proxy. Mathematically, NeurAR represents the color distribution as $\mathcal{N} (\mean, \uncert^2) $ where the RGB channels of $C$ share the same $\uncert$. Therefore, it expresses the scene function as $\nerf(\point, \dir)=( \mean, \uncert, \rho)$. Assuming colors of points on a ray and colors for different rays are independent, NeurAR yields the neural uncertainty for a viewpoint.

\be
\uncert_{v}^2=\frac 1 {\numrayset}  \sum_{r=1}^{\numrayset} \uncert_r^2=\frac 1 {\numrayset } \sum_{r=1}^{\numrayset}\sum_{i=1}^{N}W_{ri} \uncert_{ri}^2,
\eqlabel{uncertimg}
\end{equation}
where $r$ represents a camera ray tracing through a pixel, $R$ the number of rays sampled for a viewpoint, and $N$ is the number of sampled point on each ray. $W_{ri}$ is the weight of the point along the ray similar to the description in NeRF\cite{mildenhall2020nerf}. $\uncert_{ri}^2$ is the uncertainty of the color for a point learned continuously with input images added. 

At each step of the autonomous reconstruction, the reconstruction module receives a set of RGBD images associated with their viewpoints from a camera, $\{(x_i, \omega_i)\}$. With these inputs, the implicit representation $\nerf$ for the scene is updated. For the view path planning, $\uncert_{v}^2$ is calculated for sampled viewpoints.

% \subsection{MLP-A* informative planning}
% \seclabel{planner}
% Algorithm \ref{alg:vpp} shows the pseudocode of the proposed view path planning algorithm. At each step of planning, we sample views from emptyspace (line 1-4). View directions are preselected by Coarse TSDF (line 5-6) and view information gains are obtained in \eqref{view_ig} (line 7-8). We fit information gain distribution with MLP (line 9) and plan views and path by A* (line 7).

\subsection{Information Gain Field Approximation}
\seclabel{gainapprox}
% Following the greedy strategy, at each step of the autonomous reconstruction, the reconstruction module receives a set of depth and RGB images associated with their viewpoints from a camera, with which the implicit representation $\nerf$ and the neural uncertainty $\uncert_{v}^2$ for the scene are updated for 

\subsubsection{Approximation $\g$}
% NeurAR~\cite{ran2022neurar} addresses the challenges of measuring the contribution of a viewpoint to the autonomous reconstruction with implicit representations by defining the information gain of the viewpoint as the neural uncertainty in \eqref{uncertimg}, and demonstrates the significant improvement in the reconstruction quality with the planned view path guided by the uncertainty. However, inferring the neural uncertainty for a viewpoint is computational expensive. 
In \eqref{uncertimg} getting the neural uncertainty for a viewpoint is computational expensive: $R \times N $ points are passed to the MLP $\nerf$, where $R$ scales with the image size and $N$ scales with the scene size and the level of details needed by the reconstruction. In NeRF~\cite{mildenhall2020nerf}, for example, to render an image of size $800 \times 800$ given a viewpoint, $R$ is 640,000 and $N$ is set to 192. Though for the neural uncertainty in NeurAR, $R$ is set to be 1000 as a coarse approximation to save computation, the number of querying $\nerf$ is still tremendous. To improve the efficiency of the information gain query, we assume the information gain for a region is generated by a smooth continuous function and propose to
leverage a MLP ($\g$) as an implicit function approximator for the information gain field. The assumption of the smoothness is motivated by the observation that neighbouring viewpoints usually do not exhibit dramatic changes in the seen views and in their information gain. With this assumption, given a set of the information gain sampled from the field, the whole field can be interpolated. 

Specifically, at each step of the planning process, given a set of sample points $\mathcal{P}=\{ p_1,..,p_{N_{loc}}\}$ and its corresponding information gain $\mathcal{I}=\{ I_{1},...,I_{{N_{loc}}}\}$, we fit the information gain distribution by learning a mapping from a point $p$ to the information gain $I$
% \g:  $p\rightarrowtail I_p$
\begin{equation}
 \g: p\rightarrowtail I.
\eqlabel{mlp_fit}
\end{equation}
The parameters $\phi$ is optimized by defining the L2 loss between the ground truth $I$ and the estimation from $\g$, \ie $L = \sum_{i=0}^{N_{loc}} ||\g(p_i) - I_i||_2$. 

With the approximation $\g$, getting the information gain for a viewpoint is super efficient, querying $\g$ for only one time compared with querying $\nerf$ for about $R \times N $ times (\eg 100,000 times in NeuAR~\cite{ran2022neurar}). Also, $\g$ is usually much smaller than $\nerf$ in the model size and the inference is faster.

\subsubsection{View Filtering with Coarse TSDF for $\mathcal{P}$}
\seclabel{coarse_tsdf}
% By the approximation, the query efficiency during the path planning is improved by $R \times N $ times but still the information gain for a set of viewpoints with different directions in points $\mathcal{P}$ is required for the approximation. 

To further reduce the number of viewpoints required for the training of $\g$, we propose to use a coarse TSDF $T$ to filter out viewpoints. Also, as the implicit representation $\nerf$ is not efficient in querying the status of a 3D point during the path planning, an occupancy map $V$ is built up to accelerate the the planning.

$T$ represents the scene using volumetric grids by integrating partial point clouds from different viewpoints using zero-crossing method~\cite{newcombe2011kinectfusion}. From T, an occupancy map $V = V_o\cup V_e\cup V_u$ is constructed, which consists of occupied $V_o$, empty $V_e$ and unobserved $V_u$ voxels. As the volumetric maps are only for the space status query not for the fine-grained scene representation, the voxel resolution $l_{res}$ can be very large, \eg 10cm for a scene of size $3m \times 3m \times 3m$.

% The view cost is evaluated by casting a set of rays $S_r$ from a camera pose through the
% image plane to the scene, i.e.
% \begin{equation}
% C_{v} = \frac 1 {N_{ray}} \sum_{r=1}^{N_{ray}}\sigma_{r},
% \eqlabel{tsdf_cost}
% \end{equation}
% where $N_{ray}$ is the amount of sampling rays, and $\sigma_{r}$ is the uncertainty of the ray $r$. It is calculated from T as
% \begin{equation}
% \sigma_{r}= 
% \begin{cases}
% 	\left\min(1,\left\lambda_u(\frac {N_{r,u}} {N_{r}})\right^2)\right , & \ {if \ \ N_{r,o} = 0 } \\[2ex]
%     \frac 1 {1+\lambda_o d_{r}^2N_{tra}}, & \ {if \ \ N_{r,o} \ne 0  \\}
% \end{cases}
% \eqlabel{ray_cost}
% \end{equation}
% where $N_{r,o}, N_{r,u},N_{r}$ represents the number occupied voxels, unknown voxels and all voxels observed along the ray $r$ respectively, $N_{tra}$ the number of rays traversing the voxel and $d_r$ the depth of the ray $r$ (rendered from $V$). $\lambda_u$ and $\lambda_o$ are coefficients. 

The selection for the viewpoints with directions in points $\mathcal{P}$ given current viewpoint consists of three steps: 1) sampling $N_{loc}$ locations from the empty space $V_e$ within a sphere centered on the current position of the camera and its radius is $l_s$; sampling $N_{yaw}$ yaw and $N_{pitch}$ pitch angles for each location; 2) choose the best three directions according to the TSDF view cost similar to that in\cite{schmid2020efficient}; 3) choose the best direction according to \eqref{view_ig}. As the TSDF is very coarse, the computation expense of the view cost based on it is much lower than that of \eqref{view_ig}. The information gain for these viewpoints is then normalized to [0,1] to get $\mathcal{I}$.

% To filter out viewpoints, we use the view cost based on the TSDF view cost similar to that in\cite{schmid2020efficient}. 

\subsubsection{Information Gain $I$}
In NeurAR~\cite{ran2022neurar}, the target scene is in the center and the space for the view path planning is limited to a predefined ring area of a certain distance from the scene center. To free from the ad-hoc setting, we define information gain considering the distance of a viewpoint off the surface
\begin{equation}
I_{v} = 
\begin{cases}
\uncert_{v}^2,& if\ d_{min} < d_{v} < d_{max} \\
e^{-\alpha |d_{v}-d_u|}\uncert_{v}^2,& others
\end{cases}
\eqlabel{view_ig}
\end{equation}
where $d_{min}, \ d_{max}$ are minimum, maximum depth, and $d_u = (d_{min}+d_{max})/2$. $\alpha=-2/(d_{max}-d_{min})$ is decay factor. $d_{v}$ is the view depth $d_{v} = \frac 1 {|S_{r}|} \sum_{S_{r}}d_{r}$. The depth $d_r$ of a ray $r\in S_{r}$ is inferred from T. $S_r$ represent a set of rays whose depth is within the working range [$d_n$, $d_f$] of a depth camera. 

% cast from a camera pose into the scene. 

% The depth $d_r$ of a ray $r\in S_{r}$ is inferred from T. We randomly sample $N_{ray}$ rays from view $v$ and assume that the working range of the depth camera is between $d_n$ and $d_f$. The rays with the depth $d_r$ within $d_n$ and $d_f$ are selected as $S_r$.
% % \begin{equaith tion}
% % d_{v} = \frac 1 {|S_{r}|} \sum_{S_{r}}d_{r}.
% % \eqlabel{view_depth}
% % \end{equation}
% \zj{Addressed}

\subsection{Informative Path Planning}
\seclabel{Aplan}

% Though the query for the status of a volumetric representations is efficient, the ray casting and the integration inside the frustum of the viewpoint for the information gain is also computational intensive and thus hinders the application of planning algorithm requiring dense queries of the space. Due to the query inefficiency and the incomplete modelling of the information gain field,

% On the other hand, in previous view path planning work~\cite{schmid2020efficient,kompis2021informed} \yq{add more} as the computation of the information gain for viewpoints is expensive, sampling-based methods \eg RRT or RRT* are preferred in previous view path planning work. they typically require fewer number of sampled viewpoints, though many of them are not guaranteed to be optimal.

% Sampling-based methods \eg RRT or RRT* are used for the view path planning in previous work~\cite{schmid2020efficient,selin2019efficient}. However, they always generate non–optimal and zig–zagging paths~\cite{zammit2018comparison}. In the following part, we propose a novel view path planning method based the A* algorithm , which guarantees the optimality of the shortest path and the reconstruction quality.

As the computation of the information gain for viewpoints is expensive, sampling-based methods \eg RRT or RRT* are preferred in previous view path planning work~\cite{schmid2020efficient,kompis2021informed,song2018surface,isler2016information}. They typically require fewer number of sampled viewpoints, but many of them are not guaranteed to be optimal. In the following part, we propose a novel view path planning method based the A* planner , which plans shotter path and better reconstruction quality.

At each step of the planning, given the current position of the camera $p_s$, the occupancy map $V$, and a set of sampled points $\mathcal{P}$ and their information gain $\mathcal{I}$, our goal is to find a local goal node and plan an informative path to obtain higher reconstruction quality with a lower path cost. For the goal node, we choose the best viewpoint among $\mathcal{P}$ by comparing their information gain. For the informative path planning, we define a view path cost taking both the path length and the information gain of the viewpoints along the path into account,
\begin{equation}
IP(p) = f_d(p_s,p) - \lambda_{gain}  f_d(p_s,p)G(p),
\eqlabel{IP}
\end{equation}
where $f_d(p_s,p)$ is the path length between a point $p$ and the start node $p_s$, $\lambda_{gain}$ is a gain factor and $G(p)$ represents cumulative information gained along the path.  $G(p)$ is
\begin{equation}
G(p) = \frac 1 {N_{p}} \sum_{i=1}^{N_P}\g(p_i),
\eqlabel{path_ig}
\end{equation}
where $N_P$ is the number of sampling points along the path between the point $p$ and the start node $p_s$, $p_i$ is the i-th point over the path. 
$\g$ is the approximator for the information gain field. For the planning with the A* algorithm, we rank the sampling priority of points to improve sampling and planning efficiency by
\begin{equation}
Rank(p) = IP(p) + \lambda_{rank} h_d(p,p_g),
\eqlabel{rank}
\end{equation}
where $h_d(p,p_g)$ is the euclidean distance between a point $p$ and the goal node $p_g$ and $\lambda_{rank}$ is a rank factor. The step size of the planning is $l_{step}$. 

% For A* planning, we use occupied grid map from coarse TSDF and set stepsize as . 
% With an informative path, we try to select some views along the path for better reconstruction and don't introduce redundant views. Here we collect one or two views on the path.

\begin{algorithm}[!h]
    \caption{Proposed method}
    \label{alg:vpp}
    \renewcommand{\algorithmicrequire}{\textbf{Input:}}
    \renewcommand{\algorithmicensure}{\textbf{Output:}}
    
    \begin{algorithmic}[1]
        \REQUIRE Images and viewpoints$\{(x_i, \omega_i)\}$, partial reconstruction 3D models $\nerf$, $T$, current node $p_s$, sampling radius $r_s$ 
        \ENSURE Updated $\{(x_i, \omega_i)\}$, $\nerf$, $T$, $p_s$
        \STATE  $\nerf, T \gets$ Update3DScenes$(\nerf, T, \{(x_i, \omega_i)\})$          
        \STATE  $V = V_o\cup V_e\cup V_u \gets T$ 
        \STATE  $\{\omega_i\} \gets$ ViewSamplingFiltering $ (T, V_e,r_s)$ 
        \STATE  $\mathcal{P}, \mathcal{I} \gets$ InformationGain$(\nerf, T, \{\omega_i\})$
        \STATE  $\g \gets$ MLPFitting$(\mathcal{P},\mathcal{I})$
        \STATE  $p_g \gets \mathop{\arg\max}\limits_{\mathcal{P}} (\mathcal{I})$
        \STATE  $\{(x_i, \omega_i)\} \gets$ PlanExecuteViewPath$(V_e, p_s,p_g,\g)$
        \STATE  $p_s \gets p_g$
    \end{algorithmic}
\end{algorithm}
% \vspace{-5mm}
% \begin{algorithm}[!h]
%     \caption{Proposed view path planning algorithm}
%     \label{alg:vpp}
%     \renewcommand{\algorithmicrequire}{\textbf{Input:}}
%     \renewcommand{\algorithmicensure}{\textbf{Output:}}
    
%     \begin{algorithmic}[1]
%         \REQUIRE Partial reconstruction models $\nerf$ and $T$, current node $p_s$, sampling radius $r_s$ 
%         \ENSURE Updated Planned views $V_{sec}$ and path $P_{sec}$
%         \STATE  $V = \left(V_o\cup V_e\cup V_u)\right \gets T$ 
%         \STATE  $V_{sample} \gets SampleViews(V_e,r_s)$ 
%         \STATE  $V_{pre},D_{pre} \gets ViewPreselection(V_{sample},T)$ 
%         \STATE  $\mathcal{P}, \mathcal{I} \gets InformationGain(V_{pre},D_{pre},IR)$
%         \STATE  $\g \gets MLP(\mathcal{P},\mathcal{I})$
%         \STATE  $p_g \gets \mathop{\arg\max}\limits_{\mathcal{P}} (\mathcal{I})$
%         \STATE  $V_{sec},P_{sec} \gets PlanViewsPath(p_s,p_g,\g,V_e)$
%     \end{algorithmic}
% \end{algorithm}

\section{Results}

\begin{table*}[!ht]
    \vspace{-2mm}
    \centering
    \caption{Evaluations of the effectiveness and efficiency of view path for implicit neural representation.}
    \resizebox{\textwidth}{30mm}{
    \normalsize
    \setlength{\tabcolsep}{1mm}{
    % \scriptsize
    \begin{tabular}{c|cccc|cccc|cccc|cccc}
    \toprule
        % \multirow{6}*{}
         \multicolumn{1}{c}{}& \multicolumn{4}{c}{Method} & \multicolumn{4}{c}{cabin} & \multicolumn{4}{c}{childroom} & \multicolumn{4}{c}{Alexander}  \\ 
        Variant & Filter & $\g$ & IP & Planner & PSNR↑ & Acc↓ & Comp↓ & C.R.↑ & PSNR↑ & Acc↓ & Comp↓ & C.R.↑  & PSNR↑ & Acc↓ & Comp↓ & C.R.↑ \\ 
        \midrule
        
        V1\textsuperscript{\cite{ran2022neurar}} & &  &  &  RRT & 25.69 & 1.56 & 1.16 & 0.68  & 24.92  & 10.94 & 6.33 & 0.66 & 18.15 & 27.11 & 13.51 & 0.62  \\ 
        V2 & &  &  \checkmark & RRT & 26.15 & 1.37 & 1.07 & 0.74  & 26.82  & 9.73 &5.86 & 0.71 & 21.76  & 21.14 & 12.89 & 0.69      \\
        V3 & &  & \checkmark & A* & \textbf{28.67} & \textbf{1.01} & 1.02 & \textbf{0.76}  & \textbf{28.82}  & 6.03 & 4.06 & 0.77 & \textbf{24.57}  & \textbf{15.65} & 12.25 & \textbf{0.72}   \\
        V4 && \checkmark & \checkmark & RRT  & 26.98  & 1.26 & 1.06 & 0.73  & 26.21 & 8.82 & 5.63 & 0.72  & 21.48  & 21.36 & 13.03 & 0.66   \\
        V5 & & \checkmark & \checkmark & A* & 28.65  & 1.03 & \textbf{1.00} & 0.77  & 28.02  & 5.53 & 3.92 & 0.76  & 23.88 & 18.41 & \textbf{12.03} & \textbf{0.72}      \\
        V6(Ours full) & \checkmark & \checkmark & \checkmark & A* & 28.47 & 1.04 & 1.01 & \textbf{0.76}  & 28.28  & \textbf{5.44} & \textbf{3.52} & \textbf{0.78}  & 24.42 & 17.70 & 12.40 & 0.71 \\  \midrule
        
         Variant & Filter & $\g$ & IP & Planner & $N/T_{query}$  & $T_{SP}$ & $T_{GP}$ & P.L.  & $N/T_{query}$  & $T_{SP}$ & $T_{GP}$ & P.L. & $N/T_{query}$  & $T_{SP}$ & $T_{GP}$ & P.L.\\ \midrule
        V1\textsuperscript{\cite{ran2022neurar}} & &  &  &  RRT &  90 / 58  & 241 & 5109  & 44.86  &   104 / 1021  & 1723 & 52340  & 25.08 &      74 / 49  & 231 & 6153  & 430.16\\
        V2 & &  &  \checkmark & RRT &  141 / 92  & 92 & 3257 & 53.30  &  92 / 954  & 955 & 21542 & 25.30 &   119 / 75   & 75 & 2948 & 449.82 \\
        V3 & &  & \checkmark & A*  & 222 / 144   & 144 & 4932 & 42.23  & 312 / 3489   & 3490 & 75673 & \textbf{19.01} & 154 / 103  & 103 & 4538 & 409.15\\
        V4 && \checkmark & \checkmark & RRT  &  \textbf{83 / 0.06}  & 1.58 & 1781 & 46.38  & 
         \textbf{224 / 0.18}  & 2.20 & 7865 & 26.75 & 165 / 0.12   & 2.05 & 1520 & 565.31\\
        V5 & & \checkmark & \checkmark & A* & 222 / 0.15   & \textbf{1.53} & 1199 & \textbf{40.63}  &  312 / 0.22   & 1.96 & 7563 & 19.42 &  \textbf{154 / 0.10}  & \textbf{1.83} & 1293 & 411.28\\
        V6(Ours full) & \checkmark & \checkmark & \checkmark & A* & 222 / 0.15 & 1.77 & \textbf{388} & 40.98  & 312 / 0.24  & \textbf{1.94} & \textbf{1366} & 19.53 &  154 / 0.11  & 1.83 & \textbf{393} & \textbf{347.33} \\

         \midrule
    \end{tabular} 
    } 
    }
    \tablelabel{table_effect_effici}
     \vspace{-2mm}
\end{table*}

\begin{table*}[ht]
    \vspace{-2mm}
    \centering
    \caption{Evaluations of the effectiveness and efficiency with volumetric representations.}
    \resizebox{\textwidth}{10mm}{
    \normalsize
    \setlength{\tabcolsep}{1.0mm}{
    % \scriptsize
    \begin{tabular}{c|cccccc|cccccc|cccccc}
    \toprule
        % \multirow{3}*{}
        \multicolumn{1}{c}{} & \multicolumn{6}{c}{cabin} & \multicolumn{6}{c}{childroom}&\multicolumn{6}{c}{Alexander}   \\ 
        Method  & PSNR↑ & Acc↓ & Comp↓ & C.R.↑ & $T_{GP}$ & P.L. & PSNR↑ & Acc↓ & Comp↓ & C.R.↑ & $T_{GP}$ & P.L. & PSNR↑ & Acc↓ & Comp↓ & C.R.↑ & $T_{GP}$ & P.L. \\ 
        \midrule
         AEP\textsuperscript{\cite{selin2019efficient}} & 21.45 & 1.52 & 1.87 & 0.35 & 1276 & 46.23 &  14.72  & 2.17 & 12.69 & 0.70 & 4029 & 29.56 & 16.28  & 20.77 & 22.80  & 0.45 & 1319 & 677\\ 
         IPP\textsuperscript{\cite{schmid2020efficient}} & 18.74 & 1.56 & 2.03 & 0.33 & 447 & 44.27 & 9.02 & \textbf{2.06} & 67.16 & 0.42 & 1512 & 24.02 & 15.79  & 22.18 & 24.36 & 0.44 & 427 & 368 \\ 
         Our & \textbf{28.47}  & \textbf{1.04} & \textbf{1.01} & \textbf{0.76} & \textbf{388} & \textbf{40.98} &  \textbf{28.28}  & 5.44 & \textbf{3.52} & \textbf{0.78} & \textbf{1366} & \textbf{19.53} &   \textbf{24.42} & \textbf{17.70} & \textbf{12.40} & \textbf{0.71} & \textbf{347} & \textbf{393}  \\
        \midrule
    \end{tabular}
    }
    }
    \tablelabel{compare_with_tsdf}
    \vspace{-3mm}
\end{table*}

\begin{figure*}[htbp]
    %   \vspace{-2mm}
    %   \centering
      \includegraphics[width=\linewidth]{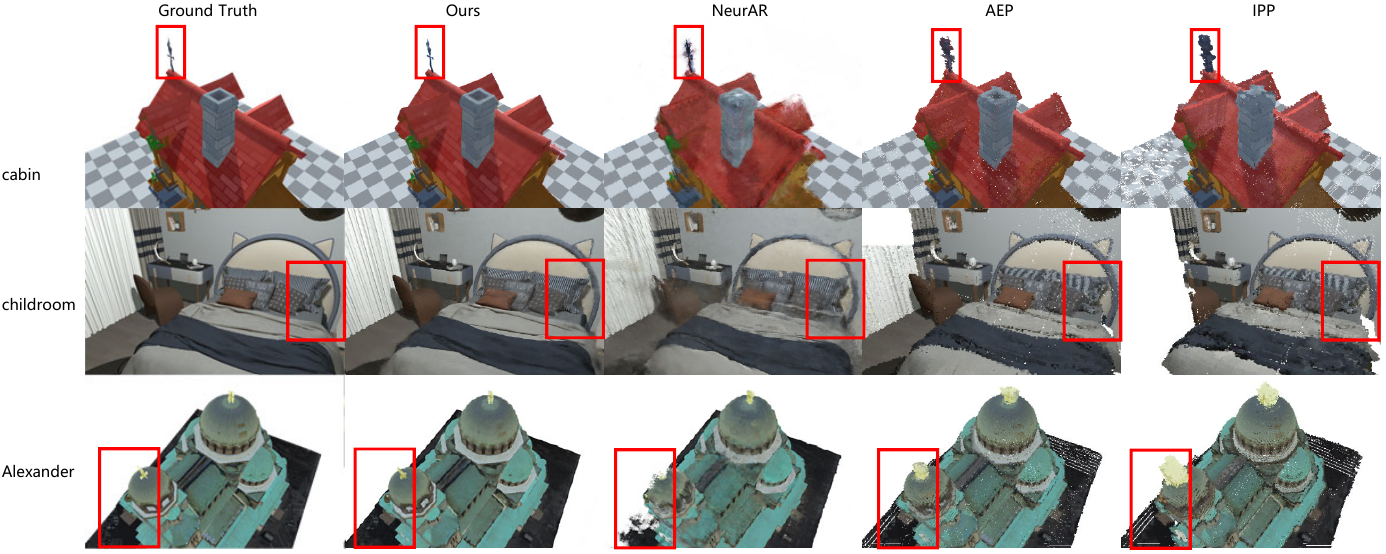}
      \caption{Comparison of the reconstruction scenes with different methods}
      \figlabel{view_compare}
\end{figure*}

\begin{figure}[htbp]
    \centering
    \includegraphics[width=\linewidth, trim={0cm 1cm 0cm 0.5cm},clip]{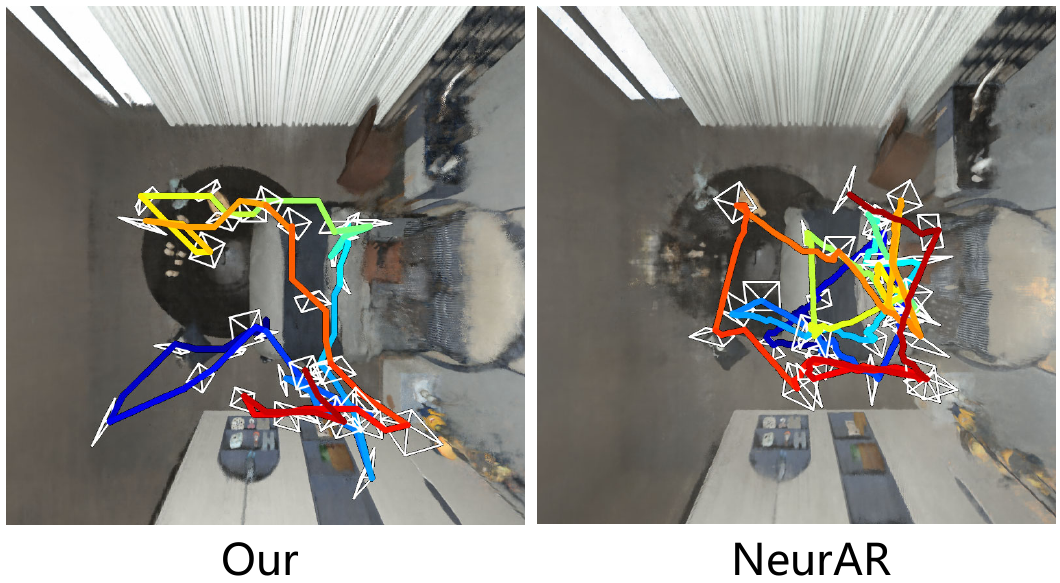}
\caption{Trajectories and the reconstruction results seen from top view. Left: Ours, Right: NeurAR ~\cite{ran2022neurar}}
\figlabel{trajectory_compare}

\end{figure}

\subsection{Implementation details}
\seclabel{impd}
Algorithm \ref{alg:vpp} summarizes the proposed efficient autonomous implicit reconstruction method. 

\subsubsection{Data} The experiments are conducted on three scenes, one small scenes $cabin$ with a size of 5m × 5m × 3m, a large scene $Alexander$ with a size of 50m × 40m × 30m, and an indoor scene $childroom$ with a size of 6m × 6m × 3m. $Alexander$ is from \cite{song2021view}, $cabin$ and $childroom$ are collected online. Similar to \cite{schmid2020efficient,ran2022neurar}, we add noise scaling approximately quadratically with depth to all rendered depth images. For $cabin$ and $childroom$, depth noise is reported from Intel Realsense L515\footnote{https://www.intelrealsense.com/lidar-camera-l515/}. For $Alexander$, depth noise is reported from Lidar VLP16\footnote{https://usermanual.wiki/Pdf/VLP16Manual.1719942037/view/}. 
% RGB image resolution is 400 × 400 for $cabin$, $childroom$ and 400 × 600 for $Alexander$.
% \zj{Addressed, add RGB image resolution here  }

\subsubsection{Implementation}  Our implicit representation is implemented based on NeurAR~\cite{ran2022neurar} and the hyper parameters are identical.  In the coarse TSDF, the near field $d_n$ is 0.5, the voxel resolution $l_{res}$ and the far field  $d_f$ are dependent on scenes. 
% The TSDF view cost \eqref{tsdf_cost} coefficients are $\lambda_u = 2.0$, $\lambda_o = 0.1$, $N_{ray}=500$. 
We set the gain factor $\lambda_d = 0.5$ in the view path cost formula \eqref{IP} and the rank factor $\lambda_{rank} = 1.5$ in \eqref{rank}. All scene-dependent parameters are listed 
in \tableref{table_param}.

\begin{table}[htbp]
    \vspace{-2mm}
    \centering
    \caption{Scene-dependent parameters.}
    \tablelabel{table_param} 
    \resizebox{75mm}{8mm}{
    \normalsize
    \setlength{\tabcolsep}{1.0mm}{
    \begin{tabular}{ccccccccc}
    \toprule
        Scene & $l_s$ & $l_{res}$ & $l_{step}$  & $d_f$ & $N_{pitch}$ & $N_{yaw}$ & $d_{min}$ & $d_{max}$  \\
         \midrule
        cabin & 3 & 0.1 & 0.2  & 6 & 3 & 5 & 2.5  & 4.5  \\
        childroom & 1 & 0.1 & 0.2  & 6 & 5 & 12 & 1.5  & 3.5   \\
        Alexander & 30 & 1  & 2  & 80 & 3 &  5 & 30  & 50  \\
        \midrule
    \end{tabular}
    }
    }
    \vspace{-3mm}
\end{table}

Our method runs on two RTX3090 GPUs similar to NeurAR~\cite{ran2022neurar}. The implicit reconstruction is on a GPU, the view path planning is on the other one. We set the maximum planned views to be 28 views for $cabin$ and $Alexander$ scenes, 40 views for $childroom$ scene. 

\subsubsection{Metric} We evaluate our method from two aspects including effectiveness and efficiency. For effectiveness, similar to iMAP~\cite{sucar2021imap} and NeurAR~\cite{ran2022neurar}, the quality of reconstructed scenes is measured in two parts: the quality of the rendered images and the quality of the geometry of the reconstructed surface. 
For the quality of the rendered images, it is measured by PSNR. For  $cabin$ and $Alexander$, 200 viewpoints are evenly distributed at the distances of 3m and 40m respectively away from the centers. For $childroom$, viewpoints are randomly sampled in the empty space and keep 1m away from the surface.
For the geometry quality, we adopt metrics from iMAP~\cite{sucar2021imap}: Accuracy (cm), Completion (cm), Completion Ratio (the percentage of points in the reconstructed mesh with Completion under 1 cm for $cabin$, scene, 5 cm for $childroom$ scene, 15 cm for $Alexander$ scene). For the geometry metrics, about 300k points are sampled from the surfaces. 

For efficiency, we evaluate the path length (meter) and the planning time (second). The total path length is $P.L.$. and the time is $T_{GP}$. For the time of our view path planning for each step of the reconstruction process, we break it into several parts for more detailed comparison: 1) the sampling time $T_{s}$ to get $\mathcal{P}$ and $\mathcal{I}$, 2) the time for the $\g$ training $T_{train}$, 3) the querying time during the planning $T_{query}$ (the number of querying points $N_{query}$ is listed along it), 4) the pure planning time without querying $T_{planner}$. We also report the total view planning time for each step $T_{SP}$ without the sampling time $T_s$, \ie $T_{SP}=T_{train}+T_{query}+T_{planner}$. $T_{train}$ for all the variants in ~\tableref{table_effect_effici} is about 1.3s to 1.8s. $T_{planner}$ for RRT and A* is less than 0.5s.

\subsection{Efficacy of the Method}

The efficacy of the method is evaluated regarding both the effectiveness and efficiency of our contributions, for which we design variants of our method based on implicit representations. 

\subsubsection {Informative Path Planning}
We make \textbf{NeurAR}~\cite{ran2022neurar} as our baseline (V1), which uses Sequential Monte Carlo (SMC) to sample views of high information gain and plans view paths by RRT. To verify the efficacy of the proposed information path planning in \seclabel{planning} compared with V1, we divide it into two components: the view path cost of \eqref{IP} and the graph based planning based on A*. Replacing the view path cost used to expand the tree in V1 with \eqref{IP} makes V2 and replacing the RRT planner in V2 with the A* planner makes V3. For these variants, the information gain for a viewpoint is computed using \eqref{uncertimg}.

The metrics of V2 in \tableref{table_effect_effici} demonstrate the proposed view path cost alone can improve the path planning in NeurAR~\cite{ran2022neurar}. When the view cost is combined with the A* algorithm (V3), the reconstruction quality shows a more significant improvement, which mainly attributes to the optimality of the A* algorithm. However, the planning with the A* algorithm typically requires more queries of the information gain, resulting in slower view planning. For example, for $childroom$ in  \tableref{table_effect_effici} the number of querying points for information gain $N_{query}$ is 312 for V2 and 92 for V3; $T_{SP}$ is 3490 seconds for V2 and 955 seconds for V3.

\subsubsection{Information Gain Field Approximation}
To verify the efficacy of the information gain field approximation by $\g$, we construct variants V4 and V5 on top of V2 and V3 respectively by querying $\g$ for the information gain instead of computing from \eqref{uncertimg}. With the approximation, the query time of the information gain $T_{query}$ is reduced by more than 10000 times, \eg 3489 seconds to 0.218 seconds for 312 queries in $childroom$ in \tableref{table_effect_effici}. The advantage of the approximation grows more significant with larger scenes or finer resolution requirement of the reconstruction. At the same time, V4 and V5 in \tableref{table_effect_effici} only see a very small drop in the reconstruction quality from V2 and V3, indicating the approximation of $\g$ is close to its ground truth. 

\subsubsection{Viewpoint Filtering with Coarse TSDF}
To evaluate the filtering using the volumetric map, we add it to V5 and make our full method V6. With the filtering, the total view planning time is reduced by about 3 times. 
% ~\tableref{table_sample_time} reports the sampling time for each step with/without the filtering using TSDF for different scenes. 

% \begin{table}[b]
%     \vspace{-2mm}
%     \centering
%     \caption{$T_s$ w/o filtering.}
%     \tablelabel{table_sample_time} 
%     \resizebox{50mm}{6mm}{
%     \normalsize
%     \setlength{\tabcolsep}{1.0mm}{
%     \begin{tabular}{cccc}
%     \toprule
%         Filter & cabin & childroom & Alexander   \\
%          \midrule
%          & 60 & 240 & 60    \\
%         \checkmark  & 18 & 50 & 18    \\
%         \bottomrule
        
%     \end{tabular}
%     }
%     }
% \end{table}

% \subsection{Extensive experiments in complex scenes}
% \seclabel{eeifs}
% To explore the scene generalization advantage of our method, we select an indoor scene $childroom$ and a large scene $Alexander$. We repeat the experiments in \secref{eeovp} and just show the results of combinations with \textbf{Pre} in \tableref{table_effect_ics} and \tableref{table_effici_ics}. We implement \textbf{NeurAR} in $childroom$ with view sampling from empty space as sampling from ball ring is not suitable for indoor scene.

% \begin{figure}[h]
%     \centering
%     \includegraphics[width=1.0\linewidth]{fig/experiments_result.png}
% \caption{experiment}
% \figlabel{experiment}
% \end{figure}

\begin{figure*}[ht]
    %   \vspace{-2mm}
    %   \centering
      \includegraphics[width=\linewidth]{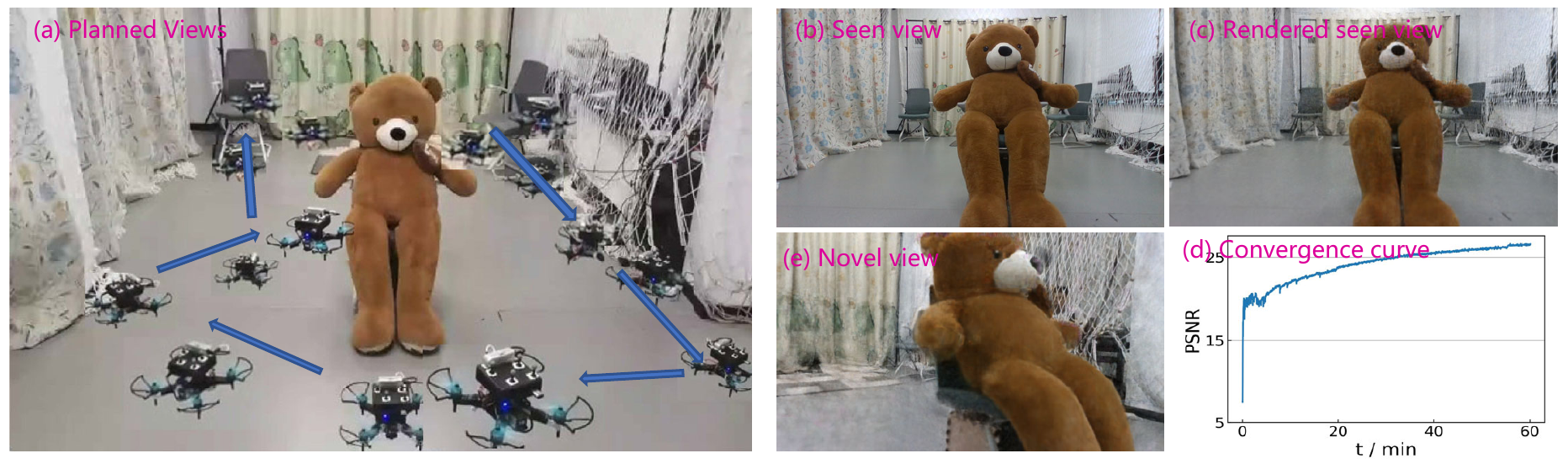}
      \caption{The experiment on a real scene.}
      \figlabel{real_scene}
       \vspace{-5mm}
\end{figure*}
% \vspace{-15mm}

\subsection{Comparison with Volumetric Representations}
\seclabel{cwovpp}

To compare with existing work on autonomous reconstruction with volumetric representations, we reimplement two methods \textbf{AEP}~\cite{selin2019efficient} and \textbf{IPP} \cite{schmid2020efficient} based on TSDF, which is one of the most used representation \cite{isler2016information,selin2019efficient,schmid2020efficient}. We set the voxel resolution of TSDF as 1cm for $cabin$ scene, 2cm for $childroom$ scene, 20cm for $Alexander$ scene. ~\tableref{compare_with_tsdf} shows that our method outperforms the reconstruction based volumetric representation in the reconstructed quality and the planning efficiency. In the childroom scene, the error measuring the accuracy of our method is higher than those of \textbf{AEP}~\cite{selin2019efficient} and \textbf{IPP}~\cite{schmid2020efficient}. This is because: \textbf{AEP}~\cite{selin2019efficient} and \textbf{IPP}~\cite{schmid2020efficient} do not fill the holes; our method using the implicit representation can interpolate the missing areas; the accuracy for the former only measures the non hole region but for the latter, the whole region. 

~\figref{view_compare} shows our method provides better reconstruction results. For more visual comparisons and results, we refer readers to the supplementary video.  ~\figref{trajectory_compare} demonstrates that the trajectory of our method expands in a larger region than that of NeurAR~\cite{ran2022neurar}.

\subsection{Ablation study}
\seclabel{ablations}
We fix the network width of $\g$ and use 4, 6, 8, 10 layer to study influence of the size on the reconstruction accuracy, shown in left of \figref{ablation_study}. We also change the number of views \figref{ablation_study} to investigate the influence. The experiment is conducted on the $Alexander$ scene. 
% Notice that $Acc$ soars with fewer viewpoints,
\begin{figure}[ht]
    \centering
    \includegraphics[width=1.0\linewidth]{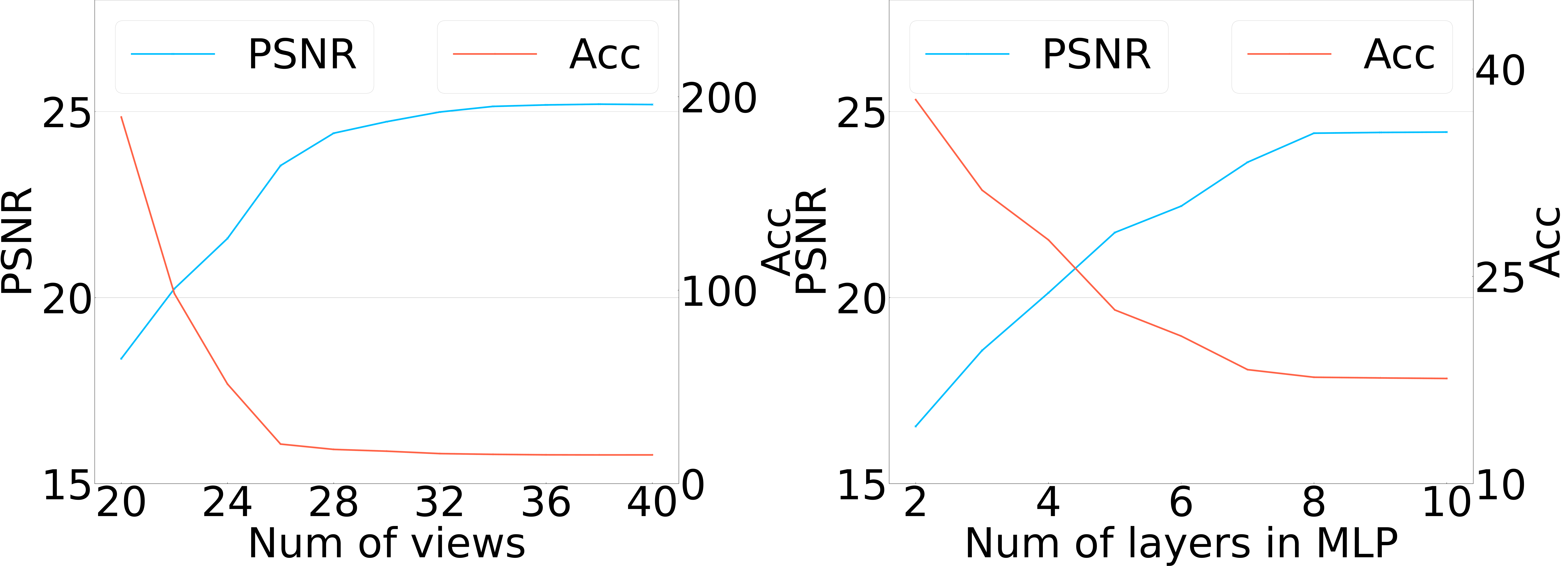}
\caption{Ablation study}
\figlabel{ablation_study}
\end{figure}

\vspace{-5mm}

\subsection{Robot experiments in real scene}
\seclabel{rers}

We deploy our proposed method on a real UAV for the reconstruction of an object placing in a room of about a size of 8m × 2.5m × 3m. The UAV is equipped with a Realsense D435i sensor\footnote{https://www.intelrealsense.com/zh-hans/depth-camera-d435i/}. The pose of the camera is provided by a Optitrack\footnote{https://optitrack.com/} system. For the scene, the UAV takes about 4 minutes to plan and executes the view path, taking 30 images. The convergence of the reconstruction module and the rendered images of the converged model are shown in in ~\figref{real_scene}.

% achieve about 20dB on the captured images at 5 minutes,  
% In \figref{real_scene} shows the reconstructed results after 4 min flight time with about 30 images. 

% and external parameters are calibrated by Iterative Closest Point (ICP). We can see

\section{Conclusion}
In the paper, we improve the efficiency of view path planning for autonomous implicit reconstruction by 1) approximating the information gain field by a function, and 2) combining the implicit representation with volumetric representations. Further, we propose a novel view path cost and plan view path with the A* planner. Extensive experiment shows that our method superiority in both efficiency and effectiveness. 

Future directions include: overcoming the limitation of the locality of our method for large scenes, tracking the camera poses with input images instead of the optitrack system, improving the reconstructed shape of the scenes by introducing more constraints, extending the reconstruction via multi-agents.

% In this paper, we have presented a coarse TSDF based efficient view path planning methods for autonomous implicit reconstruction. Coarse TSDF is used for Occupancy Grid Map, view direction preselection and adds distance constraint on Neural Uncertainty. we use \textbf{MLP} to reduce view query time, use \textbf{IP} and \textbf{A*} to improve reconstruction quality and reduce path length. Compared with other combinations for implicit representation and baselines with TSDF representation, our method demonstrates significant improvements.

% Similar to \textbf{NeurAR}, the optimization of implicit neural representation is slow, and the computation consumption is high. In the future, we will refer to some latest works such as TensoRF \cite{chen2022tensorf} to accelerate process of optimization. In addition, future works include camera pose estimation and trajectory optimization, which are very valuable for indoor and outdoor autonomous reconstruction using an UAV.

% \yq{limitation}. 

% Local based method, adding global information for large scene reconstruction. 
% multi agents.'
% Shape

% \newpage
%%%%%%%%% References
{\small
\bibliographystyle{ieeefullname}
\bibliography{view_planning}
}

\end{document}